# Exploratory Model Building

Raj Bhatnagar
ML-0008, University of Cincinnati, Cincinnati, OH 45221
bhatnagr@ucunix.san.uc.edu

## Abstract

Some instances of creative thinking require an agent to build and test hypothetical theories. Such a reasoner needs to explore the space of not only those situations that have occurred in the past, but also those that are rationally conceivable. In this paper we present a formalism for exploring the space of conceivable situation-models for those domains in which the knowledge is primarily probabilistic in nature. The formalism seeks to construct consistent, minimal, and desirable situation-descriptions by selecting suitable domain-attributes and dependency relationships from the available domain knowledge.

## 1 Introduction

In this paper we describe a formalism for exploring the space of those situation-descriptions (*also called models, scenarios, or theories*) that may be considered rationally conceivable in a domain. This exercise is the same as that of a controlled *imagination process*. Many aspects of such theory building activity, in the context of default logics, have been presented in [12]. Probabilistic knowledge derived from available databases of a domain has traditionally been used only for performing probabilistic reasoning. Our focus in this paper is to appropriately represent and use the probabilistic knowledge of a domain for performing exploratory theory-building exercises. In the following sections we first make precise our notions for the following concepts : 1. The process of imagination; 2. The form in which the probabilistic and qualitative domain knowledge are represented; 3. The requirements for a situation-description to qualify as an acceptable *imagined scenario/context*; and 4. Some types of objectives that an agent may be pursuing during the imagination process. We then discuss the computational aspects of the construction of *consistent*, *minimal* and *optimal* scenarios by an imagining agent.

## 2 The Scenario-Building Process

The scenario building (or the imagination) process has been extensively examined and discussed by many philosophers and they have examined this activity in widely varying contexts such as science, literature, philosophy, music, and painting. The scope of our formalism is limited to the imagination in the context of developing formal scientific (or other) theories.

### 2.1 A Philosophical Perspective

In his book *The Origins of Knowledge and Imagination* [1] Jacob Bronowski states *". . . every act of imagination is the discovery of likenesses between two things which were thought unlike"*. He gives the example of Newton's thinking of the likeness between a thrown apple and the moon sailing in the sky. He further states : *. . .All acts of imagination are of that kind. . . . They take the closed system, . . . open the system up, they introduce new likenesses, whether it is Shakespeare saying, "My Mistres eyes are nothing like the Sunne"* or it is Newton saying that the moon in essence is like a thrown apple. In his view the acts of *imagination* in the context of scientific discovery proceed as follows. An initial theory exists in the form of a closed system consisting of some domain attributes and some causal dependencies that inter-connect the domain attributes. An investigation is triggered by either an observation contradicting the theory or by a desire to enhance the scope of the theory by including in it more attributes from the environment of the theory's domain. The investigator then needs to discover and include in the theory a new and satisfactory causal dependency. The task of investigator's *imagination* is to provide the candidate causal dependencies and it is the role of his *critical judgment* to select one dependency from among the imagined candidates. In his lecture *Imagination and Science* [6] J. H. Van't Hoff describes this mechanism in the words : *"The so called occurring to mind results from a requisite survey of the possible cases in one's mind and a definite selection therefrom, i.e. combined efforts of imagination with the power of critical judgment are required."*

The imagination process, as described above, generates candidate causal dependencies for building up a theory. A major source from which the investigator obtains the candidate dependencies is the analogies and likenesses gathered by him from all the domains and theories known to him. A number of interesting scientific theories owe their birth to such analogy based imagination process. In [4] Donald Crosby, while discussing the evolution of Maxwell's theory of electrical and magnetic fields, states : *"Analogies with fluid flow, elastic substances, and whirling vortices had helped to bring the idea of the field into being, but once that idea had been given a firm mathematical description, these particular analogies tended to drop into the background,"* and also, *"Maxwell's equations faced backward to Newton's vision of mechanical interactions in a material medium and forward to belief in the concept of the field."* This analogy with the mechanics of fluid flow had helped Maxwell formulate the field theory which completely replaced the then prevalent and completely different notion of "actions at a distance" for explaining the effects of what we now know as electrical and magnetic fields.

In the discussions by the above quoted authors the main focus has been on incrementally correcting or extending a theory by discovering appropriate causal dependencies among the attributes of a domain. Our computational formalism follows the spirit of performing imagination by discovering appropriate causal dependencies and also generalizes the above incremental notion and seeks to hypothesize all the causal dependencies for building a hypothetical - *imagined* - theory. Since we are working with probabilistic knowledge, we equate the concept of a specific theory with that of a specific probability distribution, also called a specific *context* by us. One main consideration for the formalism is to decide on what dependencies can be extracted from probabilistic knowledge and then used as causal dependencies.

While seeking to build a hypothetical theory the imagining agent is constrained by the need to remain consistent with the body of observed evidence and is guided by a wish to make some preferred and desirable inference in the imagined theory. The *interestingness* of an imagined scenario to the agent, therefore, is determined by :

1. A set $E$ of *constraining events*. The occurrence of these events must be possible in the context of the hypothesized scenarios.
2. A *desired event*, $d$. The probability of occurrence of the event $d$ conditioned on the occurrence of the constraining events $E$ in the hypothesized context $S$, written by us as $P[d \mid E \quad (S)]$, should be very high/low (as desired by the agent).

The *interestingness* criterion for focusing the *imagination* process is based on the perception that an *imagining agent* is driven by the question — *"what are those possible contexts and scenarios in which the desired event 'd' would be very-highly/very-less likely to occur?"*, irrespective of the probability of occurrence of the *imagined context* or *scenario* itself. Having *imagined* the interesting *contexts* or scenarios, if the agent desires, he may take actions to alter the real environment in a way to make the interesting scenarios more/less likely to occur. A scientist tries to imagine those possible scenarios in which his experimental observations are explained and the *result-proposition* conjectured by him has a high probability of being true. A scenario similar to mechanical fluid flows was *imagined* to explain the observations in electrical and magnetic fields, and Maxwell's goal of having a set of equations as true in this imagined scenario is an example of this type of imagination.

A scientist, when in an exploratory mood, may not be interested in the *most reasonable* explanation of his experimental observations but may seek those, possibly less reasonable, explanations in which *his conjecture* is most likely to be true. Imagination of such scenarios would guide him towards designing those experiments and seeking those observations which would turn his imagined scenario into the most reasonable one given all the observations. This kind of imagination process is a precursor to seeking relevant evidence in the larger process of creative thinking. Such imagination process can be characterized only be a theory-building paradigm and not by a paradigm for reasoning in the context of a given theory.

### 2.2 Abduction

Abductive inference is easily understood through the following simple example given by Peirce in [11].

- The surprising fact, $C$, is observed.
- But if $A$ were true, $C$ would be a matter of course.
- Hence, there is reason to suspect that $A$ is true.

Here $C$ is the observed fact and the second sentence states the *dependency* relationship, possibly causal, (and available from the domain knowledge) that the presence of $A$ explains the presence of C. In the third statement, $A$ is an abductively inferred hypothesis. The content of the inference is the premise "If A were true, C would be a matter of course." Given a $C$, an $A$ which explains the occurrence of $C$ must be discovered by the abductive reasoner. Many $A$'s, each explaining the $C$, may exist in the reasoner's mid and he would have to choose one from all the possible candidates.

The focus of many abductive reasoners [10, 13] has been on determining the most reasonable, that is, the *parsimonious* or the *most probable* or the *least cost* [2, 14] hypotheses $A$, given the observed events $C$. The *imagining* agent is not necessarily seeking the most *reasonable* scenarios. He is guided by the *interestingness* of the candidate scenarios, while requiring only some minimal reasonableness from the hypotheses. That is, an imagining agent seeks those $A$'s in whose contexts $C$ is at least minimally explained (can possibly occur), but the *interestingness* of the hypothesis is maximum.

*Interestingness* as a preference criterion is more general than the *most reasonable* criterion. The probability of an event of interest $d$ inferred in the context of an imagined scenario is one possible criterion of interestingness which can not be simulated by using either the probability of occurrence of a scenario or the costs associated with the components of a scenario. Size of a scenario is another possible criterion of interestingness.

In the following discussion the *imagining* abductive reasoner is given the interestingness criterion in terms of the probability of occurrence of a desired event $d$, conditioned on the occurrence of events in $E$, computed in an imagined context S, and written as $P[d \mid E \quad (S)]$ by us. The agent's task is to build those hypothetical contexts $S$ in which $P[d \mid E \quad (S)]$ is very high/low irrespective of the probability of occurrence of the context $S$ itself. The contexts and the scenarios contained in them, however, must satisfy the constraints of consistency and should be minimal in size so as to exclude all irrelevant information.

## 3  Probabilistic Knowledge

For most domains the available knowledge falls into the following two categories : 1. A number of databases, each consisting of cases recorded in various contexts of the domain, and 2. Some qualitative or probabilistic domain dependencies acquired possibly from a theoretical understanding of the domain. Typically, the data part of a domain's knowledge is available in the form of a number of databases, each representing a different *context* of the domain. For example, in the domain of lung-diseases, databases may be available corresponding to different age-groups, different population groups, different specialized disease classes, and different treatment strategies. Each database represents a specialized context, reflecting the conditions under which it was recorded.

We denote by $H = (h_1, h_2, \ldots h_k)$ the set of all the known and relevant attributes for a domain. We define a *situation-description (scenario)* of a domain to be a set $T$ of attributes, along with their values such that $T \subset H$ and $T$ is non-null. By *context* we refer to a subset of some scenarios of the domain. In the lung-disease example, each individual database mentioned above represents a specialized *context* of the domain of lung diseases, and approximates the joint probability distribution only for that *context*.

One way in which a database of cases serves as a source for domain-knowledge is that we can extract from it the important inter-attribute dependencies. These dependencies may take the form of conditional probability functions or the *likelihoods* used in Bayesian inference. Let us say that the set $P$ contains sufficient number of conditional probability functions (dependency relationships) derived from a database having $T$ as its set of attributes so that the pair $< T, P >$ approximates the probability distribution reflected by the database. A different pair $< T, P >$ for each contextual database represents the probability distribution approximated by its corresponding database. An example of dependencies learnt from a contextual database $S$ is the set of probabilities $P[symptom_i \mid disease_j \quad (S)]$. Bayesian reasoning systems treat these likelihood dependencies as the *invariants* of the *context* $S$ for performing probabilistic inference. That is, the conditional probability $P[symptom_i \mid disease_j \quad (S)]$ remains constant in the context $S$, and is used as a dependency relation between the probabilities of the *symptom* and the *disease* events. The problem of compacting a joint probability distribution into the set $P$ of important dependency relationships has been extensively researched and some of the popular graph based representation methods have been presented in [3, 7, 8, 10].

### 3.0.1  Abduction with Probabilistic Knowledge

One type of abductive reasoning in a context $S : < T, P >$ has been described in [8, 10] as follows. A set of observed attributes corresponding to attributes $E_1, E_2, \ldots E_k$, (contained in set $E$ where $E \subset T$) are given. That is, the instantiations $E_i = e_{ij}$ have been fixed by the observations. The set $A = (T - E) = (A_1, A_2 \ldots, A_m)$ represents all the unobserved attributes of the context under consideration. The objective of the abductive exercise is to find an assignment for each unobserved attribute $A_i$ such that :

$$P[A_1 = a_{1i}, A_2 = a_{2j} \ldots A_m = a_{mn} \mid E_1 = e_{1i}, \ldots$$
$$E_k = e_{kl} \quad (S)] \qquad (1)$$

is maximum among all possible sets of assignments to the attributes in set $A$. We say that $A^h$ refers to that set of assignments which results in the maximum value for the above probability. the content of the abductive exercise (in a manner parallel to Peirce's formulation) can be summarized as follows :

- In context $S$ the surprising events of the set $E$ are observed.
- But if assignments $A^h$ were true, $E$ would be a matter of course.
- In the context $S$ there is no other set of assignments for the attributes of set $A$ which is more likely to occur along with the occurrence of the observed events contained in set $E$.
- Hence, there is reason to suspect that $A^h$ is true about the observed situation.

The third statement above says that the $A^h$ selected is the most probable set of assignments, given the observed evidence. This type of reasoning is in the spirit of inductive reasoning where we seek the most reasonable hypothesis given the observed events. The *imagining* agent needs to perform a more widespread exploration directed at the most interesting scenarios and not restricted to only the most reasonable explanations of the observations.

### 3.1 Probabilistic Knowledge - "Imagination" Perspective

One basic way in which the imagining agent's view of the probabilistic knowledge and reasoning differs from the traditional reasoning methods is as follows. The traditional methods consider a domain's probability distribution as a unified whole within which all reasoning activity is performed. The context invariant conditional probability functions is only a way of representing the complete distribution. We consider an imagining agent who possesses the probability distribution and the causal dependency ordering of all the attributes in the distribution. The latter knowledge may be available from a deeper understanding of the domain to which the database pertains. He orders the attributes according to their causal (may be partial) order and determines the context invariant dependencies such that in the dependency $P[a \mid B \quad (S)]$ $B$ is a set of attributes which are strictly causal predecessors of event $a$. For the imagination agent these causal context invariant dependencies is all the knowledge that is needed.

For the imagining agent the domain knowledge is represented by $n$ contextual databases, $< T_1, P_1 >, < T_2, P_2 > \ldots \ldots < T_n, P_n >$ where the distribution $< T_i, P_i >$ represents the $i^{th}$ context $S_i$ for the domain. It is possible that an attribute $h \in H$ may appear in more than one context and in such a case there may exist a dependency of the type $P[h \mid a.subset.of.T_i \quad (S_i)]$ in the set $P_i$ of every context $S_i$ that includes $h$.

The various $P_i$'s, therefore, represent the *invariant* dependencies for different *contexts* and are useful for probabilistic inference only within their respective contexts.

#### 3.1.1 Difference in Perspectives

The main difference between the imagination agent and the traditional probabilistic inference is the following :

- For traditional reasoning the only meaningful knowledge is a probability distribution $< T, P >$ within which the instances of reasoning are performed.

- The imagining agent considers each *invariant* causal dependency, learnt from a particular context, as a basic entity of the domain knowledge. As discussed in the section on philosophical perspective of imagination, the imagining agent feels free to use causal likenesses learnt from one context to build and test other interesting hypothetical contexts. The context invariant causal dependencies, therefore, are the basic units which the imagination agent considers as models of contextual causal phenomena that can be used as building blocks for building the descriptions of other hypothetical contexts.

The *contextual* causal dependencies learnt from individual contexts are reflective of the underlying causal processes for the context and therefore are viewed as possible causal building blocks from which other, possibly not yet encountered, contexts may be constructed. The causal dependencies learnt from a complete domain's probability distribution can be seen as weighted accumulations of the causal dependencies of individual contextual databases. Therefore, an instance of abductive reasoning with probabilistic knowledge in the traditional sense would not be following the process of selecting some causal dependencies and leaving out the other. It would be aggregating over all the known causal dependencies that affect any particular attribute.

The analogical thought as practiced by the imagining agent is similar to the capacity of a thinker to visualize a concept in a hypothetical context, different from the one in which it was learnt. In the context of probabilistic knowledge, the *invariant* dependencies characterizing a context are placed in a different hypothetical *context* by the imagining agent to imagine the descriptions of hypothetical scenarios. This capability is, arguably, the foundation of imagination process as discussed earlier, which in turn is the foundation of creativity. Theory building using defaults and default-logic has been presented in [12] and in our case the causal context invariant dependencies may be seen as analogous to the defaults which may or may not be applicable in any particular situation.

When we mix and match the contextual dependencies learnt from disparate contexts to hypothesize new contexts we don't have enough information to determine the probability of occurrence of this new concocted context. This is because the marginals represented by the contextual databases are not sufficient to construct the complete joint distribution from which the probability of occurrence of a particular context may be determined. Since our formulation of an imagining agent, as explained in the previous section, does not need to determine the probabilities of occurrence of various scenarios, the above shortcoming is not of much significance.

The abduction exercise that the *imagining agent* needs to perform can be stated as follows.

- The surprising events contained in the set $E$ are observed.

- But if the context were to be $S^h$, represented by $< T_h, P_h >$ then $E$ would be a matter of course.

- There is no other context $S$ in which $P[d \mid E\ (S)]$ is greater than what is obtained in the context $S^h$.

- Hence, there is reason to suspect that $S^h$ is the most interesting context and the various scenarios obtained from $S^h$ (by making attribute assignments) are the most interesting scenarios.

The main difference between this abduction formulation and the one described in section 3.0.1 can be seen in terms of the second and the third statements of the abduction formulation. The imagining agent is seeking to hypothesize that context in which interesting inferences can be made and the traditional reasoner is trying to hypothesize the most probable scenario in a given context. The computational problem of the traditional reasoner's abduction, as presented in [8, 10], is to find the most suitable assignment $A^h$ in a given context $S$. The corresponding problem for the imagining agent's abduction task is to construct the context $S^h$ by selecting an appropriate subset $T_h$ from the set of domain attributes $H$ and an appropriate subset $P_h$ from the set of all the known contextual invariant dependencies, such that the probability of occurrence of $d$ in $S^h$ meets the interestingness criterion.

## 4  The Dependency Relation

For each contextual database of a domain, we can represent the joint probability distribution by the pair $< T, P >$ where the set $P$ may contain either all the Chow dependencies [3], or all the Bayesian dependencies [8, 10], or all the qualitative dependencies known from a theoretical understanding of the domain.

We define a relation $D$ for the complete domain, including information from all its contextual databases, such that

$$D \subseteq \bigcup_{i=1}^{n-1} \{H^i \times H\} \text{ where } H^i = H \times H \times \ldots \times H,\ i\ times$$
(2)

where $H$ is the set of all the attributes in all the contextual databases, $domain(D) = \bigcup_{i=1}^{n-1} \{H^i\}$ and $range(D) = H$. This relation has the same character as the set $P$ of a joint probability distribution. The difference is that $D$ contains all the dependencies learnt from all the contextual databases or the qualitative knowledge of the domain and it does not necessarily constitute a consistent description of a joint probability distribution.

To construct the dependency relation $D$ for the complete domain we repeat the following with each available contextual database. We first order (or partial order, if sufficient knowledge is not available) the attributes of the database in such a way that each attribute is followed in the sequence by only those that can possibly causally influence it. Given this ordering of attributes, we determine the sets of Bayesian network dependencies, and the known qualitative dependencies for the database. A union of all these sets is the contribution of this contextual database to the dependency relation $D$ of the complete domain. The relation $D$ therefore contains all the dependencies derived from each available contextual database. From the perspective of traditional probabilistic reasoning $D$ is an unnatural and meaningless medley of conditional probability functions but for the imagining agent the relation $D$ is the source from which to derive the candidate causal dependencies to build a hypothetical theory. The issue of maintaining some consistency has been addressed in a later section.

For the imagining agent the elements of set $D$ are independent entities in the sense that a particular *dependency* can be used without worrying about the probabilistic dependence of its consequent attribute on some other attributes of the context not included in the *dependency*. This is because in a Bayesian *dependency* a consequent node is in fact probabilistically independent given the antecedent attributes of the *dependency*. Therefore, by using these dependencies the agent is not making any assumption about any probabilistic independences. The independence among dependencies is a characteristic of the way the Bayesian or other dependencies are constructed.

## 5  Structure of an Imagined Context

The task of the imagining agent is to hypothesize a context $S^h$ such that $P[d \mid E\ (S^h)]$ is in accordance with the interestingness criterion. The interestingness criterion may seek this probability value to be minimum, maximum, or satisfy some specified constraint. The description of the context $S^h$ consists of the pair $< T_h, P_h >$, specifying *some* joint probability distribution. It is an arbitrary, *imagined* distribution but must be consistent as a description of a distribution. Imagining a *context*, therefore, is the same as constructing a relevant, complete and consistently specified approximation of a probability distribution by using the known dependencies as building blocks. An analogy with building consistent theories using defaults in default logic [12] can be drawn here. Our context building task is accomplished by selecting an appropriate set of dependencies from the domain dependency relation $D$. However, any arbitrary choice for the set $P_h$ may not be an acceptable description of a context. The constraints that a hypothesized context $S^h$ must follow are the following:

1. Sufficiency : A hypothesized context $< T_h, P_h >$ is considered sufficient if all the constraining attributes $(E)$ and the event of interest $d$ are included in the set $T_h$. The objective of sufficiency criterion is to ensure that in the imagined contexts the probability of occurrence for each constraining event can be computed. Only those hypothesized contexts are of interest to the imagining agent in which scenarios with non-zero probabilities for all the constraining events can be constructed.

2. Consistency : A hypothesized context is considered consistent if it is a complete and consistent description of a joint probability distribution - even though a completely hypothetical joint probability distribution.

3. Minimality : A hypothesized context is considered minimal if it contains only those attributes and dependencies that are needed to show the possibility of occurrence of the constraining events and the high/low probability of the occurrence of the desired event.

## 5.1  A Sufficiently Large Context

A hypothesized context that is sufficiently large in description should : 1. Include the attribute $d$ corresponding to the event of interest. 2. Include the constraining attributes contained in the set $E$. 3. Include a subset $R$ of the dependency relation $D$ such that their associated conditional probability functions define a complete joint probability distribution. 4. Be an *Explanation* for the occurrence of the desired event $d$ and the constraining events $E$.

In the abduction example given by Peirce in [11] the abductive hypothesis $A$ is considered an explanation of the observed event $C$. In this same spirit, w our notion of an explanation is as follows :

### 5.1.1  A *Context* as an Explanation

Each conditional probability function in $D$, say derived from a Bayesian network, $P[node.n \mid parents.of(node.n) \quad (S)]$ represents a dependency of the context represented by the Bayesian network. The dependencies selected for constructing a hypothesized context can be arranged in a graphical network to show the unique Bayesian Network that they represent. We define a *path* in the graphical representation of a context as follows :

**Definition** : A path from a node (attribute) $v_1$ to a node $v_k$ in a context $<T, R>$ is defined as a sequence $\{v_1, w_1, v_2, w_2 \ldots w_{k-1}, v_k\}$, where each $w_i \in R$, each $v_i \in T$, and for each $w_j$

1. either $v_{j+1}$ or $v_j$ is its consequent node and the other is its antecedent node;

2. no $v_j$ is the consequent node of both $w_{j-1}$ and $w_j$; and

3. no $v_i$ or $w_i$ is repeated in the sequence;

4. $v_k$ is not an antecedent node of any $w_i$ included in the sequence.

(end-definition)

The conditions in the above definition ensure that each path : 1. is a connected sequence of causal dependencies; 2. is non-cyclical; and 3. connects vertices $v_1$ and $v_k$ without containing a vertex which is a causal descendent of both $v_1$ and $v_k$. A context description can also be viewed as a collection of various such paths.

We consider Peirce's description of abduction again and replace in it the fact $A$ above by a path of dependencies, the observed event $C$ by a pair of attributes, and the explanation $A$ by a the path. The resulting reformulation of Peirce's example can be stated as follows:

- The surprising events, $e_i$ and $e_j$ are observed to occur

- If the path of dependencies $A$ between these two observed events were to be active in a context then $e_i$ and $e_j$ would have occurred as a matter of course.

- Hence, there is reason to suspect that the path of causal dependencies $A$ is active in the situation.

The content of the above inference is the premise "If the path of dependencies $A$ were active, the events $e_i$ and $e_j$ would occur as a matter of course." The path $A$, therefore, is a possible explanation for the occurrence of the two events.

The above definition of an explanation restricts the contexts that can be hypothesized by the imagining agent. The agent is restricted to explain each constraining event in relation to another constraining event, or the event of interest. A single attribute, with a single dependency containing the attribute as its consequent node, may also be considered as a possible explanation of an event corresponding to the attribute. This latter definition, however, results in a large number of trivial explanations in which no constraining event is connected either to another constraining event or to the desired attribute. Imposing the above, more restrictive, definition of an explanation forces the imagining agent towards those imagined *contexts* in which the attributes are relatively much more connected to each other. The *contexts* with the more liberal definition of an explanation are also valid from the *imagination* perspective but are more disconnected as hypotheses and add more complexity to the computational process of constructing the interesting *contexts*.

### 5.1.2  Contents of a Context

A *sufficiently large* context $S^h$ is one in which the posterior probability $P[d \mid E \quad (S^h)]$ and the probabilities $P[(e_i \in E) \quad (S^h)]$ can be computed and it is an explanation in the above described sense of the constraining and the desired events.

We say each element $d_i$ of $D$ has the form $(x, y)$ where $y$ is the consequent attribute of $d_i$ and $x$ is the set of antecedent attributes of $d_i$. We define the functions $conseq(d_i)$, $antec(d_i)$, $conseq^*(r)$, and $antec^*(r)$ as follows:

- $conseq(d_i)$ is the attribute $y$ in the pair $(x, y)$ of $d_i$.

- $antec(d_i)$ is the set of attributes $x$ in the pair $(x, y)$ of $d_i$.

- $conseq^*(R)$, where $R \subseteq D$, is the set of attributes containing $conseq(d_i)$ for each $d_i \in R$, and contains no other attributes.

- $antec^*(R)$, where $R \subseteq D$, is the set of attributes containing $antec(d_i)$ for each $d_i \in R$, and contains no other attributes.

A *Sufficient* description of a context is the pair $< T, R >$ such that :

1. $T \subseteq H$, $E \subseteq T$, $d \in T$
2. $R \subseteq D$,
3. $d \in conseq^*(R)$
4. $E \subset conseq^*(R)$
5. There is a path from each $e_i$ to either the attribute $d$ or another constraining attribute $e_j$, and there is a path from $d$ to at least one $e_j$.

The third and the fourth conditions stated above imply that a hypothesized context must include dependencies that have $d$ and all the members of $E$ as consequent nodes. That is, we must include in the hypothetical context those dependencies which can be viewed as the causes for the occurrence of $d$ and the members of $E$. The fifth condition states that in a sufficiently large context each event, in conjunction with some other event of interest has been explained. This condition would cause many nodes, other than $d$ and the constraining events, to be included in set $T$.

A sufficiently large pair $< T, R >$ selected as above would be an acceptable *context* only if the dependencies in $R$ constitute a consistent description of a joint probability distribution.

## 5.2 A Consistent Context

The notion of consistency of a hypothesized *context* is derived from the perspective of a *context* being the same as a probability distribution. That is, a context $< T, R >$ is *consistent* only if the dependencies in $R$ completely describe a joint probability distribution for the attributes in $T$. This consistency condition ensures that the resulting context is neither under-specified nor over-specified. Using the chain rule and deleting those conditioning attributes which are not a part of the dependency relationship, the joint probability distribution for the attributes of the set $T$ can be written as :

$$P[t_1, t_2, ... t_m] = \quad P[t_1 \mid G_1] * P[t_2 \mid G_2] * ... \\ * P[t_{m-1} \mid G_{m-1}] * P[t_m] \quad where (3)$$

$G_i \subseteq \{t_{i+1}, t_{i+2}, \ldots, t_m\}$. The similarity between the above expansion and a *context* being hypothesized is that each product term on the right hand side is a dependency in $R$ which helps constitute the hypothetical *context*. Keeping in mind such an expansion for a probability distribution, we should construct a context $< T, R >$ by including in $R$ dependencies from the domain dependency set $D$ such that :

1. there is *one* and only *one* dependency in $R$ corresponding to each product term of equation 3;
2. each product term and the corresponding dependency in $R$ have exactly the same attributes and in the same relative positions; and
3. set $R$ includes no other dependencies;

A pair $< T, R >$ that follows the above correspondence completely and consistently describes *some* hypothetical joint probability distribution without either under or over specifying it.

For some given $d$ and $E$ it is possible to select a number of different subsets $R$ from the set $D$ such that each of these choices consistently describes a joint probability distribution. The objective of the imagining agent is to prefer those contexts in which the probability of the desired event is maximum/minimum (as desired).

## 5.3 The Minimal Context

A hypothesized context $< T, R >$ is considered minimal if the following are true :

1. Removal of any dependency $r \in R$ from the context would disrupt a path between either attribute $d$ and an attribute $e_i \in E$, or between attributes $e_i \in E$ and $e_j \in E$.
2. Every $t \in T$ is included in at least one $r \in R$.

A path is the abductively hypothesized explanation for all the consequent nodes included in the path. The above definition of minimality implies that we don't want any such dependencies in the context that are not playing a role in explaining the occurrence of either the attributes $d$ or a constraining attributes. It should, however, be noted that a *context* in the above defined senses of sufficiency, consistency, and minimality can be a set of disjoint directed acyclic graphs. Each Bayesian Network is a directed acyclic graph [10] if we assume a direction in each dependency from the antecedent nodes to the consequent node.

With the constraints on an acceptable Context specified, we now examine the computational framework for constructing the contexts in which scenarios with desired probabilities values for the events of interest can be constructed.

## 6 Constructing Preferred Contexts

The computational task of constructing the optimal contexts is formulated by us as a state-space search guided by an admissible heuristic function. The AI search algorithm $A^*$ given in [9] has been directly employed in our implementation. Each *state* is a partially constructed context $<t, r>$ where $t \subset T$ and $r \subset R$ of some completely described context $<T, R>$. In the start state the set $t$ contains the attribute $d$ and all the constraining attributes contained in the set $E$.

To maintain the minimality and the consistency of the resulting context-description, a dependency $d \in D$ can be added to the set $r$ of a state $<t, r>$ only if the following hold :

- $conseq(d) \in (t \cup antec^*(r))$, and
- $conseq(d) \notin conseq^*(r)$.

That is, only that dependency can possibly be added to a state whose consequent node already exists in the state and there is no other dependency in the state with the same consequent node. The successor states of a search state are all those states that are obtained by adding one dependency to the parent state. It can be seen with little reasoning that each possible context description that is consistent and minimal can be built using the above incremental step of adding one dependency at a time. Also, all context descriptions that are built using the above constrained incremental process are consistent. The minimality is guaranteed by stopping the incremental steps as soon a sufficiently large context has been created.

The goal state $S$ for the search algorithm is that completely described context in which each constraining attribute $e_i$ is connected by a *path* to either the attribute $d$ or another constraining attribute $e_j$. That is, the the sufficiency requirements for a context as discussed in section 5.1 are satisfied.

Tha $A^*$ search algorithm can yield the best, the second best, etc. hypotheses until all possible hypothetical contexts have been generated. The capability to output multiply hypothetical contexts, rank ordered according to their interestingness, is a very desirable aspect of an imagining agent. Some computational formalisms [10] can only compute the best or the best and the second best hypotheses and this would be an undesirable restriction on an imagining agent.

The next important aspect of the $A^*$ search algorithm is the heuristic function used for ordering the partially constructed contexts. If the *Merit Function* whose value must be maximized by the desired context is given by

$$F(S) = (P[d = d_l \mid E\ (S)]) \quad (4)$$

then in order for the search to be admissible, we need an estimating function $W(ss)$ such that

- $ss$ is a partially constructed context, and
- the value $W(ss) \geq F(S)$ where $S$ is any complete *context* that can be constructed by adding more dependencies to $ss$.

With an admissible estimating function we are guaranteed to get the optimal *context* as its first output from the $A^*$ search. Further *contexts*, in order of decreasing merit value can also be obtained by continuing the search process.

For the *Merit function* $F(S)$ specified above, an estimating function $W(ss)$ can be specified as follows:

$$W(ss) = (\max_{i,j,k...n} P[d = d_l \mid a, b, c, \ldots g \quad (h)]) \quad (5)$$

where $h$ is that dependency which has $d$ as its consequent node, and $a$, $b$, $c$ etc. are those antecedent attributes of $h$ that lie on a path which can possibly exist in an extended version of the state $ss$. We can show by simple arithmetic that $W(ss)$ is indeed an upper bound on $F(S)$ as required above.

**Proposition :** Considering that $S$ refers to a completed *context*, and $ss$ is a partially constructed context of $S$, the estimating function $W(ss)$ as described above computes a value that is not less than $P[d = d_l \mid E\ (S)]$ for all those $S$ that can be generated from $ss$ by addition of more dependencies.

**Proof** : Since $S$ is a consistent context, the attribute $d$ in it can be the consequent node of only one dependency (according to the conditions of consistency of a context). Let us say this dependency is $h$ and $a$, $b$, $c \ldots g$ etc. are its antecedent nodes. If $E$ is the set of constraining nodes included in $S$ then we can say that

$$P[d = d_l \mid E\ (S)] = \sum_{j,k..l} (P[d = d_l \mid a_j \wedge b_k \wedge ..g_l\ (S)]$$
$$* P[a_j \wedge b_k \wedge ..g_l \mid E\ (S)]). \quad (6)$$

The first product term on the right hand side above represents the conditional probability values for the dependency $h$ and the second term represents the probability computed for the antecedent attributes based on the constraining events. The right hand side of the above expression can be rewritten as :

$$\sum_{j,k..l}(P[d = d_l \mid a_j \wedge b_k \wedge ..g_l\ (h)] * P[a_j \wedge b_k \wedge ..g_l \mid E\ (S)]). \quad (7)$$

This expression can be viewed as

$$\sum_r (a_r * x_r)$$

where $a_i$'s correspond to the first terms and $x_i$'s to the second terms of the product. Since the sum of the probability values for all the events in a consistent context must be *one*, we can say that

$$\sum_i x_i = 1.$$

Considering this constraint, we can say that

$$max(\sum_i a_i * x_i) = \max_i(a_i). \quad (8)$$

That is, for the Merit function under consideration we can say that

$$max(P[d = d_l \mid E \quad (S)]) =$$
$$\max_{j,k...l}(P[d = d_l \mid a_j \wedge b_k \wedge \ldots g_l \quad (h)]). \quad (9)$$

(End-Proof)

The above described search process would therefore stop when a pair $< T, R >$ has been constructed in which the probability of occurrence of the event $d = d_l$ is the maximum among all possible *contexts* that can be constructed given the constraining events, desired event, and the storehouse of dependencies, the relation $D$, from which the imagining agent can select the causal dependencies.

The selected merit function is a very simple interestingness criterion and the suggested estimating function is also a very weak upper bound on the merit function. More intelligent estimating functions, that are stronger upper bounds on $F(S)$ can be designed. The imagining agent can also attempt variations of the merit functions to achieve different types of objectives for the imagination exercise.

The imagining agent, thus, can produce hypothetical situation descriptions that are sufficient, consistent, and minimal for an imagination task. The imagined contexts need not be possible in the real world but they are only conceivable by an imagining mind. The actions that should be taken based on the imagination of a conceivable situation descriptions is the next step after the imagination task.

## 7 Conclusion

Intelligent systems have tended to use the probabilistic knowledge of a domain for traditional probabilistic reasoning alone. The task of constructing the conceivable contexts and scenarios of a domain requires that we move away from the *most probable*, the *most likely*, or the *least cost* focus of the inductive probabilistic inference. We have presented in this paper a formalism for using the probabilistic knowledge available in a domain, by methods different from those of traditional probabilistic reasoning. We have defined the nature of sufficient, consistent, and minimal hypothetical contexts that can be conceived by an imagining agent and have briefly outlined an AI search based computational formalism for arriving at the interesting imagined hypotheses. We have presented justifications of this formalism from the perspective of an *imagination* process and have highlighted the differences between an imagination formalism and a reasoning formalism using the probabilistic knowledge.

**Acknowledgment**

This research was supported by National Science Foundation Grant Number IRI-9308868. This is a smaller version of a complete report which can be obtained from the author. We are greatful to the reviewers for providing very useful comments on the earlier version of this paper.

## References

[1] Jacob Bronowski. *The Origins of Knowledge and Imagination*, Yale University Press, 1978.

[2] Eugene Charniak and Solomon E. Shimony. Probabilistic Semantics for Cost Based Abduction, *Proceedings of the 1990 National Conference on Artificial Intelligence*. pp 106-111.

[3] C. K. Chow and C. N. Liu. Approximating Discrete Probability Distributions with Dependence Trees. *IEEE Transactions on Information Theory*, vol. IT-14, No. 3, May 1968

[4] Donald A. Crosby and Ron G. Williams. Creative Problem Solving in Physics, Philosophy, and Painting : Three case Studies, *Creativity and the Imagination* edited by Mark Amsler, University of Delaware Press, 1987, pp 168-214.

[5] Johan de Kleer and Brian C. Williams. Diagnosis With Behavioral Modes, *Proceedings of the IJCAI, 1989* Morgan Kaufmann, pp 1324-1330.

[6] J. H. Van't Hoff. *Imagination in Science*, Springer Verlag New York Inc., 1967.

[7] S. L. Lauritzen and D. J. Spiegelhalter, Local Computations with Probabilities on Graphical Structures and their Application to Expert Systems, *The Journal of Royal Statistical Society, Series B*, vol. 50, No. 2, pp 157-224, 1988.

[8] Richard E. Neapolitan *Probabilistic Reasoning in Expert Systems*, John Wiley and Sons, Inc. 1990.

[9] Nils Nilsson. *Principles of Artificial Intelligence*, Tioga Press, 1980.

[10] J. Pearl. *Probabilistic Reasoning in Expert Systems : Networks of Plausible Inference*. Morgan Kaufmann, San Mateo, CA, 1988.

[11] Charles S. Peirce. The Philosophy of Peirce - Selected Writings, Edited by Justus Buchler, Harcourt, Brace and Company, 1940.

[12] David Poole. A Logical Framework for Default Reasoning, *Artificial Intelligence*, vol. 36, pp 27-47, 1988.

[13] Yun Peng and James A. Reggia. Abductive Inference Models for Diagnostic Problem-Solving. Springer Verlag 1990.

[14] Eugene Santos Jr. On the Generation of Alternative Explanations with Implications for Belief Revision. *Proceedings of the Seventh Conference on Uncertainty in Artificial Intelligence*, pp 339-347.